\documentclass{article}
\usepackage{cite}
\usepackage{amsmath,amssymb,amsfonts}
\usepackage{algorithm,algorithmic}
\usepackage{graphicx}
\usepackage{subfig}
\usepackage{caption}
\usepackage{textcomp}
\usepackage{xcolor}
\usepackage{multirow}
\usepackage{booktabs}
\usepackage{authblk}
\usepackage{lscape}
\def\BibTeX{{\rm B\kern-.05em{\sc i\kern-.025em b}\kern-.08em
    T\kern-.1667em\lower.7ex\hbox{E}\kern-.125emX}}

\DeclareMathOperator*{\argmin}{arg\,min}
\DeclareMathOperator{\prox}{prox}

\newcommand\blfootnote[1]{%
  \begingroup
  \renewcommand\thefootnote{}\footnote{#1}%
  \addtocounter{footnote}{-1}%
  \endgroup
}

\title{Continuous Learned Primal Dual\\}
\author[1]{Christina Runkel}
\author[]{Ander Biguri}
\author[]{Carola-Bibiane Sch\"{o}nlieb}
\affil[]{Department of Applied Mathematics and Theoretical Physics, University of Cambridge}

\begin{document}

\maketitle

\blfootnote{Preprint.}
\begin{abstract}
Neural ordinary differential equations (Neural ODEs) propose the idea that a sequence of layers in a neural network is just a discretisation of an ODE, and thus can instead be directly modelled by a parameterised ODE. This idea has had resounding success in the deep learning literature, with direct or indirect influence in many state of the art ideas, such as diffusion models or time dependant models. Recently, a continuous version of the U-net architecture has been proposed, showing increased performance over its discrete counterpart in many imaging applications and wrapped with theoretical guarantees around its performance and robustness. In this work, we explore the use of Neural ODEs for learned inverse problems, in particular with the well-known Learned Primal Dual algorithm, and apply it to computed tomography (CT) reconstruction. 
\end{abstract}

\section{Introduction}
Computed Tomography (CT) is an ubiquitous imaging technique in modern medicine that allows for imaging of patients using X-rays. In brief, CT relies on measuring a series of images corresponding to the attenuation of X-rays by the object of interest (a human, in medicine), by rotating the X-ray source and detector around the patient, typically around a full circle. CT reconstruction thus refers to the problem of obtaining the image that produced the measurements, often called sinograms.

This problem, mathematically modelled by the Radon transform (line integrals over a domain), is \textit{ill-posed}, as, in general, breaks the three conditions that define a \textit{well-posed} problem: Firstly the solution is not continuously dependant on the measurement, as small changes in the measured sinogram will represent large changes in the image. Secondly it has no unique solution in the general sense, particularly in undersampled or limited view tomography. Finally, under sufficient measurement noise, there may be no solution. This theoretical analysis has direct implication in medicine, as signal noise is directly related to X-ray dose, which ideally is desired to be reduced as much as possible, as X-rays are ionizing-radiation, which leads to cell dead and can increase likelihood of cancer arising in living tissue. Similarly, the non-uniqueness can be an issue, as albeit most standard CT is performed using full circular trajectories, clinical applications like breast tomosynthesis or image guided surgery often have physical limitations on the scanning range of the CT machine, and thus inherently cannot acquire all the required data to ensure uniqueness of solutions. 

In practice, it is thus rare to reduce the noise and the scanning range, as the reconstruction is often unusable. But, if a robust enough reconstruction method can be found, dose reduction becomes feasible. Classically (and very often, clinically) the method that solves the CT reconstruction is the Filtered Backprojection (FBP) algorithm, an approximation of the inverse of the aforementioned Radon transform. As this method assumes a continuous sampling with no noise, it performs sufficiently well under those conditions, but rapidly degrades with increased noise and undersampling.

Other methods have been proposed, based on the variational regularization approach\cite{scherzer2009variational}
\cite{benning_burger_2018} that, by using the physics of CT, can iteratively solve the CT reconstruction problem, generally with much better performance against noise, particularly under appropriate choices of regularization, such as Total Variation. In recent years, these methods have been enhanced by using data-driven methods, i.e. Machine Learning (ML). 

A variety of methods have been proposed for data driven CT reconstructions, but in this work we will focus on the Learned Primal Dual (LPD). The goal of this work is a robustness enhancement of learned methods, and showcasing a proof of concept using LPD. The motivation of this work is driven by Neural Ordinary Differential Equations (Neural ODEs), a way to interpret the typical convolutions and layers that convolutions neural networks (CNNs) are made of as a discretization of a continuous ODE. This continuation of the discrete layers produces better performing networks that are provably robust to noise, and have been shown to outperform their discrete counterparts in practical scenarios (see e.g., \cite{pinckaers2019neural, cheng2023continuous}). Given that noise rejection is a key feature of a good CT reconstruction solver, this work proposes to put together data-driven models and Neural ODEs, to further enhance their performance. We propose the Continous LPD (cLPD), an idea that however is feasible to implement in any other data-driven inverse problem, in principle.

\section{Methods}

In this section we first introduce CT reconstruction, then the LPD algorithm and Neural ODEs. This leads to the novelty in this work, the cLPD and its architecture. 

\subsection{Variational formulation of reconstruction}

Mathematically, one can write the CT reconstruction problem as seeking to recover a function (the image) $x:\mathbb{R}^3\rightarrow\mathbb{R}$ from the measured projections, described by the Radon transform as $y(\ell)=\int_{\ell}x(z)\,\mathrm{d}z, \ell\in\mathcal{L}$, where $\mathcal{L}$ represents the lines in $\mathbb{R}^3$ from the X-ray source to each detector, defined by the scanner geometry and rotation. This is often linearized and discretized as
\begin{equation}
    Ax = y +\tilde{e} \label{eq:linear}
\end{equation}
where $A$ represents the integral over the lines (often referred as the \textit{forward operator}), $x$ is a vector representing the pixel values, $y$ is a vector representing the measured sinogram values and $\tilde{e}$ is the noise or error, either from measurement of from the linearization. To solve \ref{eq:linear} in a robust manner, the variational regularization approach has found significant success in the literature, proposing the following optimization:

\begin{equation}
\hat{x} = \argmin_x \mathcal{D}(y,Ax) + \mathcal{R}(x)\label{eq:vr}   
\end{equation}
where $\mathcal{D}$ measures the data fidelity between the measurement and the image estimate (most commonly the $l^2$ distance in CT, due to the characteristics of the noise) and $\mathcal{R}$ is a regularization function that promotes images of desired properties, also called a \textit{prior}. 

The optimization literature has proposed many methods to solve \ref{eq:vr}, given particular choices of $\mathcal{D}$ and $\mathcal{R}$. These methods have been shown to outperform the FBP algorithm under most conditions, given appropriate choice of functions and parameters. 

\subsection{Data-driven methods: Learned Primal Dual}

In recent years, NN have been proposed to solve problems like CT in \ref{eq:linear}. While many methods can be proposed as a \textit{post-processing} of a reconstruction, generally FBP as 
\begin{equation}
    \bar{x} = \mathcal{N}_\theta(\hat{x}) \label{eq:NN}
\end{equation}
being $\mathcal{N}_\theta$ a NN parametrized by $\theta$. While these produce solutions $\bar{x}$ of high quality, they solutions are not guaranteed to be of small $\mathcal{D}(y,A\bar{x})$ (i.e. fitting to the measured data), as there is no such constraint in $\mathcal{N}_\theta$. Thus \textit{data-driven model-based} methods where proposed in the literature, attempting to mix data driven methods with algorithms that use explicit use of the physics knowledge of the model, $A$. While several methods exist, in this work we focus on the  LPD\cite{lpd_tmi}.

LPD was formulated starting from the Primal Dual Hybrid Gradient (PDHG) algorithm\cite{cham_pock}, that solves \ref{eq:vr} using classical methods, and can be expressed as in algorithm \ref{algo:PDHG},

 \begin{algorithm}
 \caption{Primal Dual Hybrid Gradient}\label{algo:PDHG}
 \begin{algorithmic}[1]
 \renewcommand{\algorithmicrequire}{\textbf{Input:}}
 \renewcommand{\algorithmicensure}{\textbf{Output:}}
 \REQUIRE $x_0$, $z_0$, $\sigma >0, \tau >0$, $\rho\in[0,1]$
  \FOR {$i = 1, ...$}
  \STATE $z_{i+1} \gets \prox_{\sigma\mathcal{D}}(z_{i}+\sigma A\bar{x}_{i})$
  \STATE $x_{i+1} \gets \prox_{\tau\mathcal{R}}(x_{i}-\tau A^Tz_{i+1})$
  \STATE $\bar{x}_{i+1} \gets x_{i+1} +\rho(x_{i+1} -x_{i}) $
  \ENDFOR
 \end{algorithmic} 
 \end{algorithm}
with an appropriate initialization of $x_0$ (e.g. FBP) and $z_0$ (often zero). This algorithm uses \textit{proximal} operators, defined as
\begin{equation}
    \prox_{\tau\mathcal{F}}(x) = \argmin_{u} \mathcal{F}(u)+\frac{\tau}{2}\|u-x\|_2^2.
\end{equation}

LPD thus proposes to replace these proximal operators, and also the update step for $\bar x_{i+1}$ for NNs, leading to:

 \begin{algorithm}
 \caption{Learned Primal Dual}\label{algo:LPD}
 \begin{algorithmic}[1]
 \renewcommand{\algorithmicrequire}{\textbf{Input:}}
 \renewcommand{\algorithmicensure}{\textbf{Output:}}
 \REQUIRE $x_0$, $z_0$
  \FOR {$i = 1, ..., I$}
  \STATE $z_{i+1} \gets \Gamma_{\theta_i^d}(z_{i},A\bar{x}_{i},y)$
  \STATE $x_{i+1} \gets \Lambda_{\theta_i^p} (x_{i},A^Tz_{i+1})$
  \ENDFOR
 \end{algorithmic} 
 \end{algorithm}

In algorithm \ref{algo:LPD}, the number of iterations $I$ is predefined (therefore the common name of \textit{unrolled method}), and networks $\Gamma_{\theta_i^d}$ and $\Lambda_{\theta_i^p}$ therefore are defined by a different set of parameters $\theta_i$ in each iteration. In practice often $z_{i+1}$ and $x_{i+1}$ are composed of several channels, but only one of them is used to update the respective variable.
Interestingly, these \textit{primal} and \textit{dual} networks require small parametrizations, as the intuition of replacing a proximal suggest, they do not need to represent a complex transform, only a small step change. In comparison to a typical NN, LPD uses the operator $A$, thus limiting the results to the space of valid images. It has been shown that in simulated studies, LPD outperforms most well known classical variational methods and post-processing NN methods of the form of \ref{eq:NN}, leading to many variations being proposed\cite{eldar_spm,fista_net,itnet}. It is important to note that while LPD has the form of a classical optimizer with convergence guarantees, such properties are lost once parametrized with a network\cite{mukherjee2023learned}. It is more appropriate to see the entirety of algorithm \ref{algo:LPD} as a single network $LPD_\theta(y)$

The LPD is finally trained given a set of training data $T_j=(x^j,y^j), j\in[1,J]$ with a loss function $L(\theta)$ minimizing the empirical loss
\begin{equation}
    L(\theta) = \frac{1}{J}\sum_{j=0}^J\|LPD_\theta(y^j) - x^j\|,
\end{equation}
and using the resulting $\theta$, employing typical minimization algorithms from machine learning literature, such as Adam. 

For the purpose of this work, however, we are interested in how $\Gamma_{\theta_i^d}$ and $\Lambda_{\theta_i^p}$ are constructed. As its standard in imaging applications, these are constructed as a series of discrete convolutions. While this method of constructing NNs is overwhelmingly the standard, there is evidence that one can obtain better results if these convolutions are modelled by a continuous function, rather than a discrete operation. This continuous representation was proposed and named Neural Ordinary Differential Equations or, Neural ODEs \cite{chen2018neural}.

\subsection{Neural Ordinary Differential Equations}
Neural ordinary differential equations (NeuralODEs) as introduced in \cite{chen2018neural} are based on the fact that neural networks like ResNet \cite{he2016deep} can be seen as an Euler discretisation of a continuous transformation \cite{haber2017stable, ruthotto2020deep, lu2018beyond}
Every discrete layer thus computes $x_{t+1} = x_t + f_{\theta_t}(x_t)$ for a parametrised function $f_{\theta_t}$ and an input $x_t$. By reducing the size of the steps, i.e., adding more layers to the network, in the limit the network $f_{\theta}$ describes the dynamics of hidden units as the following ordinary differential equation (ODE):
\begin{equation}
    \frac{\partial x(t)}{\partial t} = f_{\theta}(x(t), t).
\end{equation}
The output of the network $x(T)$ thus can be computed by solving the ODE initial value problem at time $T$ via standard ODE solvers. Computing the backward step to compute gradients during training of the network requires backpropagating through the solver. As this is memory-inefficient due to the solver possibly needing hundret of function evaluations, Chen et al \cite{chen2018neural} introduced the adjoint method. The adjoint method treats the solver as a black box and uses a second ODE going backward in time, starting with the gradients of the original output with respect to the loss function. Using automatic differentiation, the gradients with respect to the parameters can be calculated in a memory efficient way. 

Neural ODEs are known to be memory and parameter efficient and robust to noise while providing theoretical underpinnings from the theory of ordinary differential equations. 
\subsection{The Continuous Learned Primal Dual}
The aim of the continuous learned primal dual algorithm (cLPD) is to combine both the advantages of the classical learned primal dual algorithm with those of neural ODEs. Continuous learned primal dual therefore replaces the discrete convolutional blocks in both networks $\Gamma_{\theta_i^d}$ and $\Lambda_{\theta_i^p}$ by continuous neural ODE blocks $\Gamma_{\theta_i^d}^c$ and $\Lambda_{\theta_i^p}^c$ (see Algorithm \ref{algo:cLPD}). 
As neural ODEs have proven to be more robust to noise, a better handling of noise that is inherent in the data can be achieved - a feature that is particularly useful for CT reconstruction.
 \begin{algorithm}[h]
 \caption{Continuous Learned Primal Dual}\label{algo:cLPD}
 \begin{algorithmic}[1]
 \renewcommand{\algorithmicrequire}{\textbf{Input:}}
 \renewcommand{\algorithmicensure}{\textbf{Output:}}
 \REQUIRE $x_0$, $z_0$
  \FOR {$i = 1, ..., I$}
  \STATE $z_{i+1} \gets \Gamma^c_{\theta_i^d}(z_{i},A\bar{x}_{i},y)$
  \STATE $x_{i+1} \gets \Lambda^c_{\theta_i^p} (x_{i},A^Tz_{i+1})$
  \ENDFOR
 \end{algorithmic} 
 \end{algorithm}
\subsection{Network Architecture}
\begin{figure}[th]
    \centering
    \subfloat[Dual iterates, $\Gamma_{\theta_i^d}$.]{\includegraphics[width=0.45\textwidth]{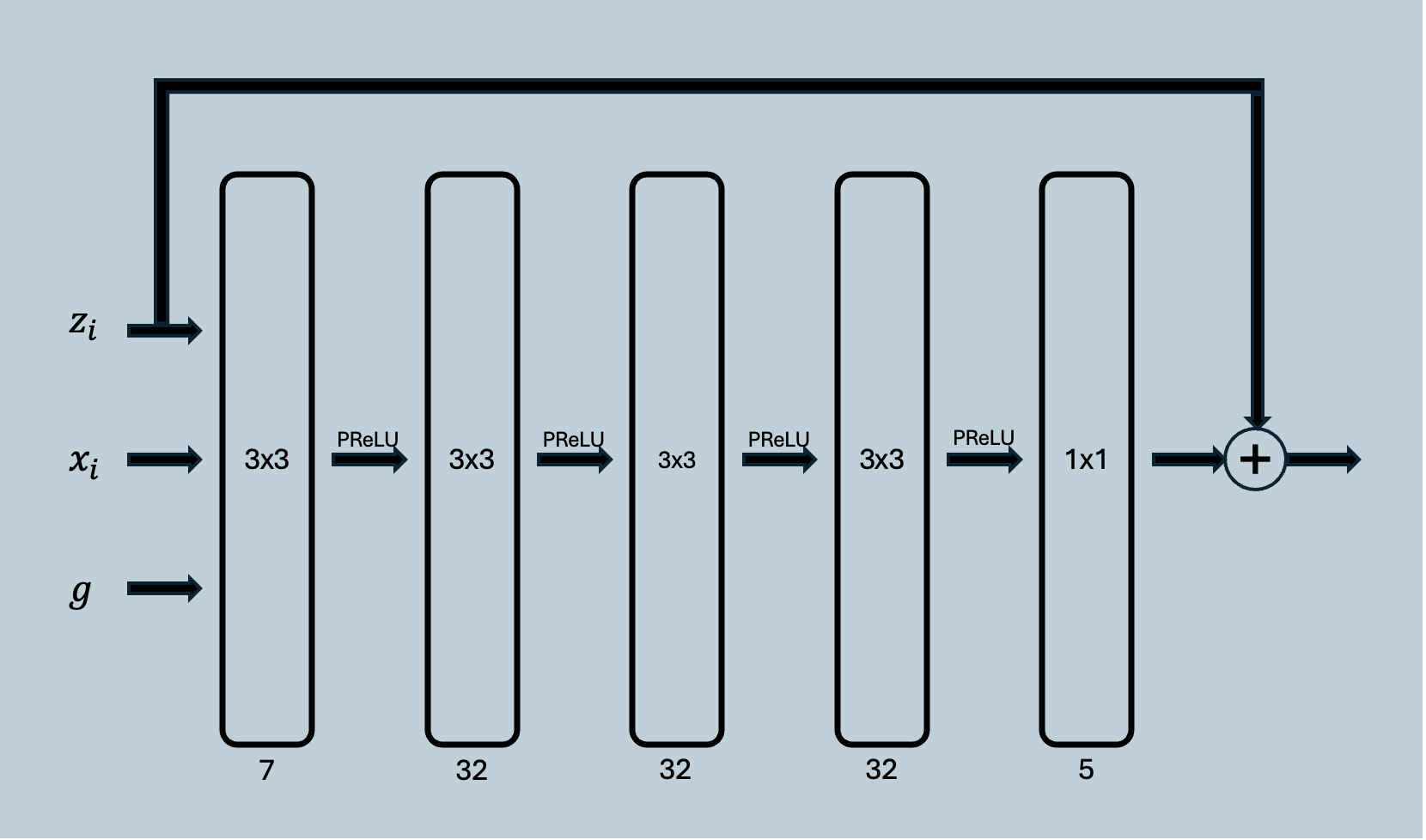}\label{fig:cLPD_dual}}
    \hfill
    \subfloat[Primal iterates, $\Lambda_{\theta_i^p}$.]{\includegraphics[width=0.45\textwidth]{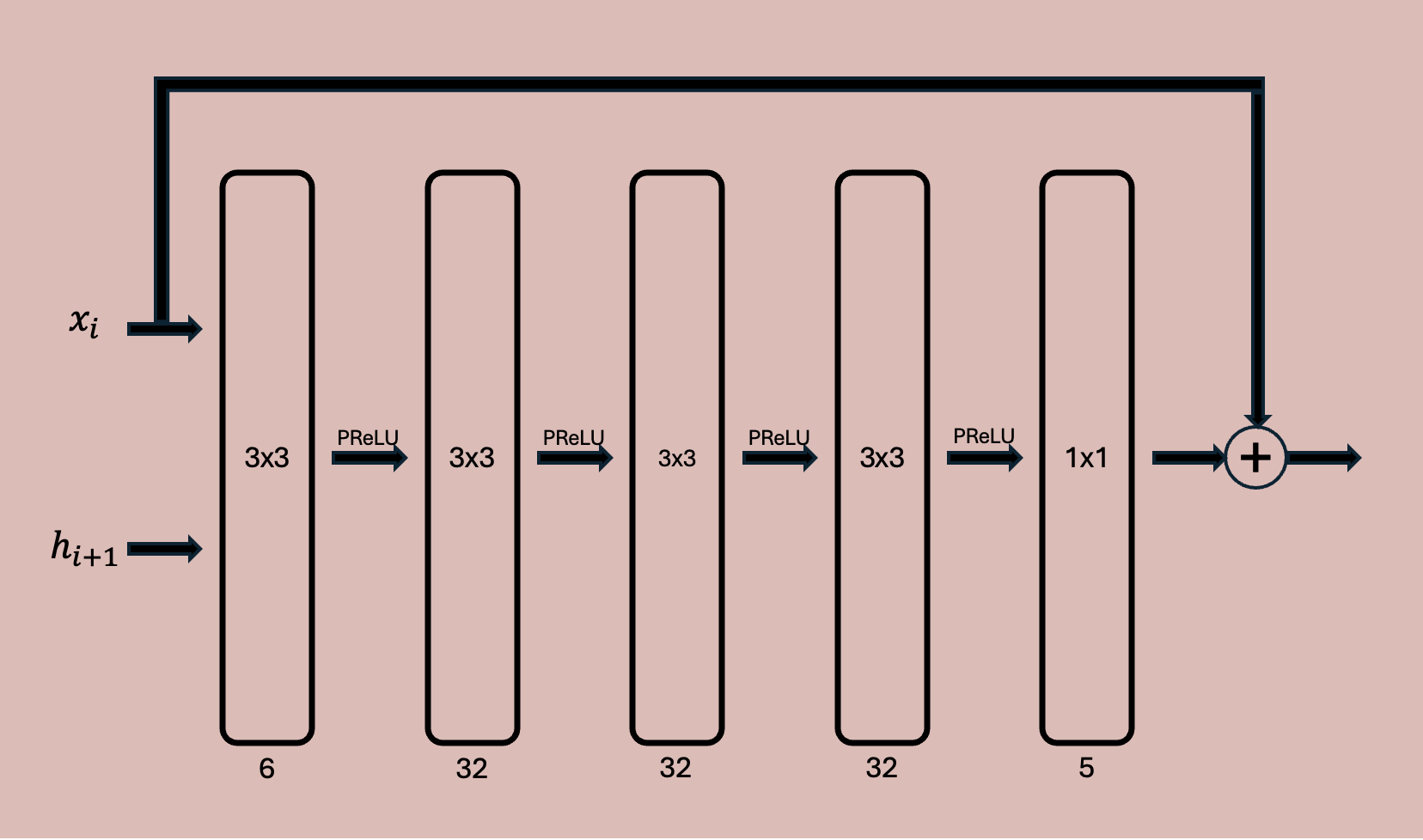}\label{fig:cLPD_primal}}
    \caption{Network architecture for both the dual and primal iterates of the (continuous) learned primal dual algorithm. Each of the rectangles describes a convolution and ODE for the LPD and cLPD, respectively. The number of input channels is denoted below the box and the kernel size specified in the middle of the rectangle.}
    \label{fig:network_architecture}
\end{figure}

The network architecture for both the dual and primal iterates of the continuous learned primal dual algorithm is highlighted in Figure \ref{fig:network_architecture}. We define the ODE by using five convolutional layers with parametric ReLU (PReLU) activation functions for primal and dual iterates.

\section{Experimental Setup}\label{sec:experimental_setup}
To emphasise the advantages of our continuous learned primal dual algorithm, we conduct experiments on the following different radiation doses and geometries:\newline
\textbf{1. Clinical setting}: We firstly test the clinical setting, i.e., a clinical radiation dose on a full circle.\newline
\textbf{2. Reduced dose setting}: An ongoing challenge in CT reconstruction is minimising the radiation dose per patient. This can be achieved by either reducing the X-ray dose or decreasing the number of angles that get measured. We thus test the following experimental settings:
    \begin{enumerate}
        \item[a)] \textit{Extreme low dose, full circle}: Reducing the X-ray dose by measuring over the full circle.
        \item[b)] \textit{Sparse angle, clinical dose}: Reducing the number of angles to measure while keeping the clinical X-ray dose.
        \item[c)] \textit{Sparse angle, extreme low dose}: Reducing both the number of angles to measure and the X-ray dose.
    \end{enumerate}
\textbf{3. Restricted setting}: Clinicians additionally are also interested in a restricted setting. In this setting, it is not possible to measure the full circle but just up to a very limited angle - increasing the difficulty of reconstructing images drastically.
    \begin{enumerate}
        \item[a)] \textit{Limited angle, clinical dose}: We firstly test the restricted setting on a clinical X-ray dose.
        \item[b)] \textit{Limited angle, extreme low dose}: Additionally, we then try the limited angle setting on an extreme low X-ray dose.
    \end{enumerate}
In the following, we will analyse the results for the experimental settings above for our continuous learned primal dual algorithm, the standard learned primal dual with discrete layers and filtered backprojection as comparison to a classical method. We train both the cLPD and LPD with a batch size of $2$, learning rate of $10^{-4}$ and the original LPD parameters used in \cite{lpd_tmi} for $100$ epochs on the LIDC-IDRI dataset \cite{lidc_dataset} using the Adam optimiser \cite{kingma2014adam}.
\section{Experimental results}
\begin{figure*}[th]
    \centering
    \subfloat[Visual results for clinical setting (1.).]{\centering \includegraphics[width=0.7\textwidth, height=0.4\textheight, keepaspectratio,trim={90 100 60 100},clip]{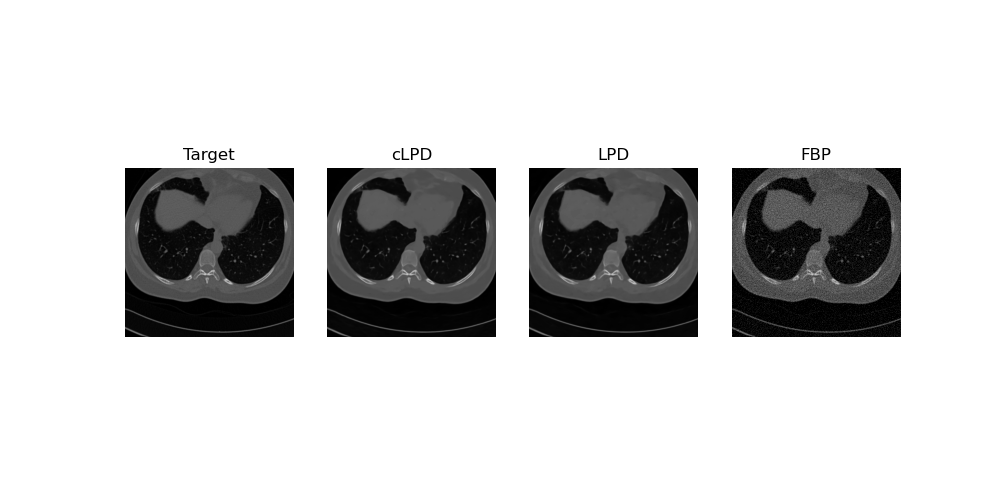}\label{fig:full_low}}
    \hspace{2cm}
    \subfloat[Visual results for extreme low dose, full circle setting (2.a)).]{\centering \includegraphics[width=0.7\textwidth, height=0.4\textheight, keepaspectratio,trim={90 100 60 100},clip]{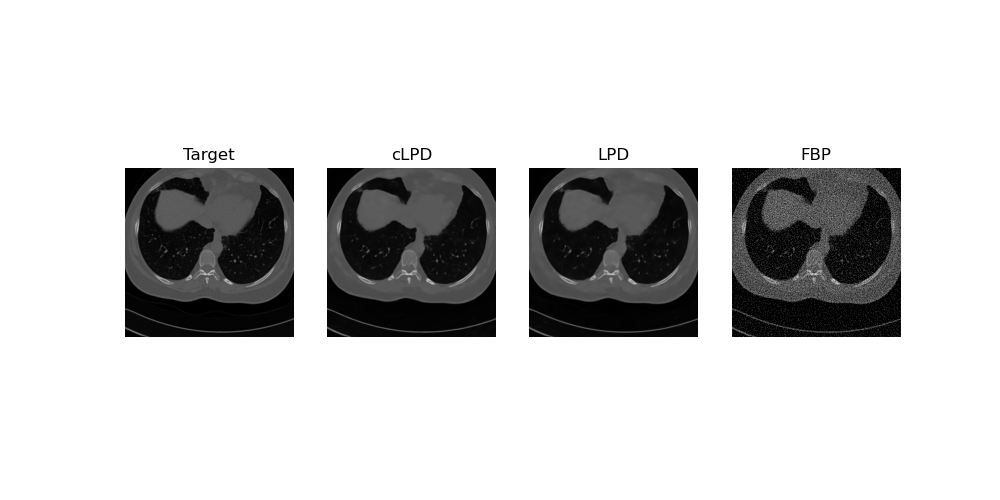}\label{fig:full_extreme_low}}
    \hspace{2cm}
    \subfloat[Visual results for sparse angle, clinical dose setting (2.b)).]{\centering \includegraphics[width=0.7\textwidth, height=0.4\textheight, keepaspectratio,trim={90 100 60 100},clip]{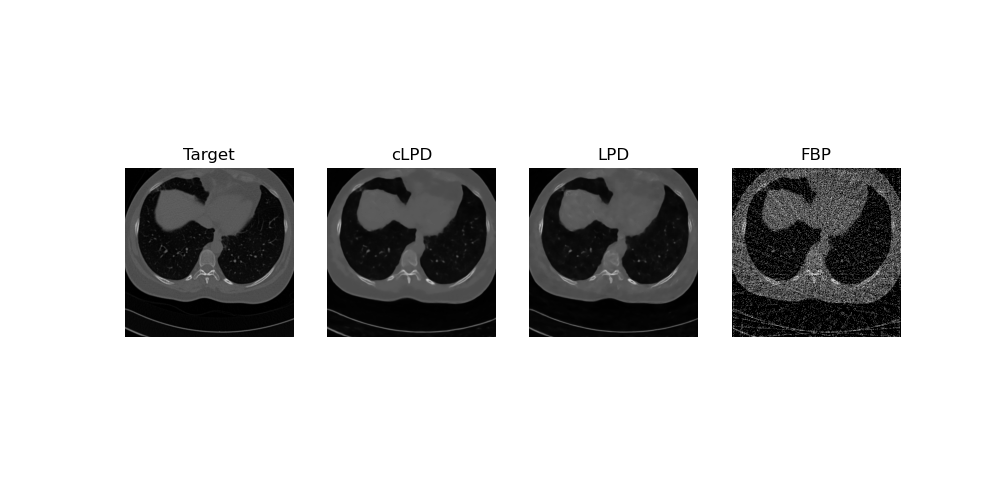}\label{fig:sparse_low}}
    \hspace{2cm}
    \subfloat[Visual results for sparse angle, extreme low dose setting (2.c)).]{\centering \includegraphics[width=0.7\textwidth, height=0.4\textheight, keepaspectratio,trim={90 100 60 100},clip]{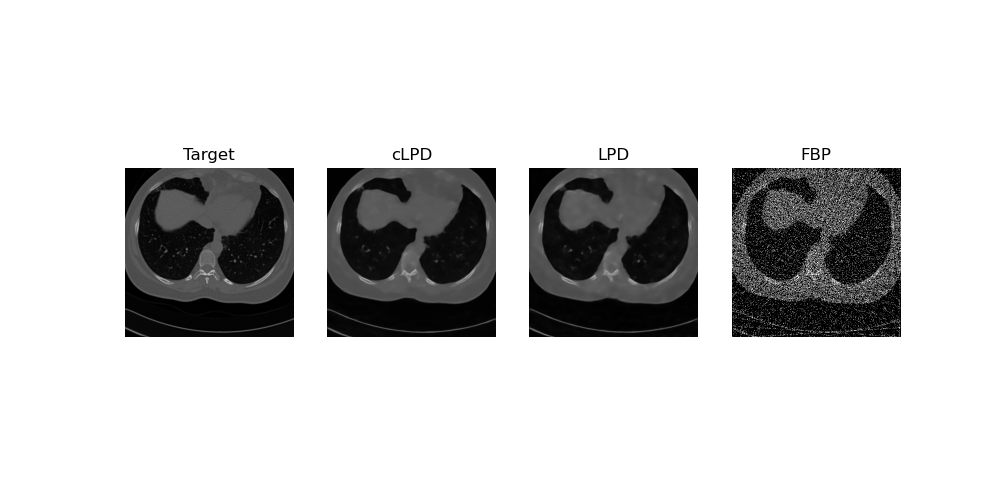}\label{fig:sparse_extreme_low}}
    \hspace{2cm}
    \subfloat[Visual results for limited angle, clinical dose setting (3.a)).]{\centering \includegraphics[width=0.7\textwidth, height=0.4\textheight, keepaspectratio,trim={90 100 60 100},clip]{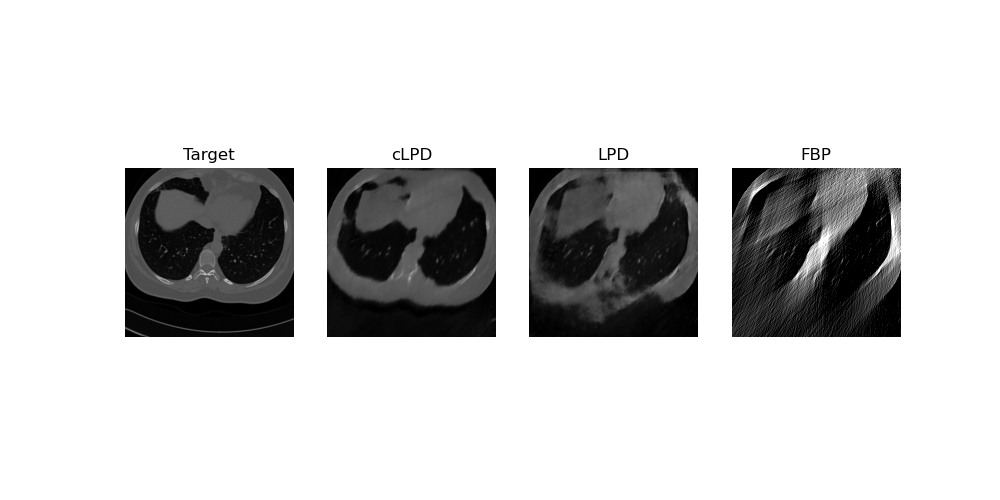}\label{fig:limited_low}}
    \hspace{2cm}
    \subfloat[Visual results for limited angle, extreme low dose setting (3.b)).]{\centering \includegraphics[width=0.7\textwidth, height=0.4\textheight, keepaspectratio,trim={90 100 60 100},clip]{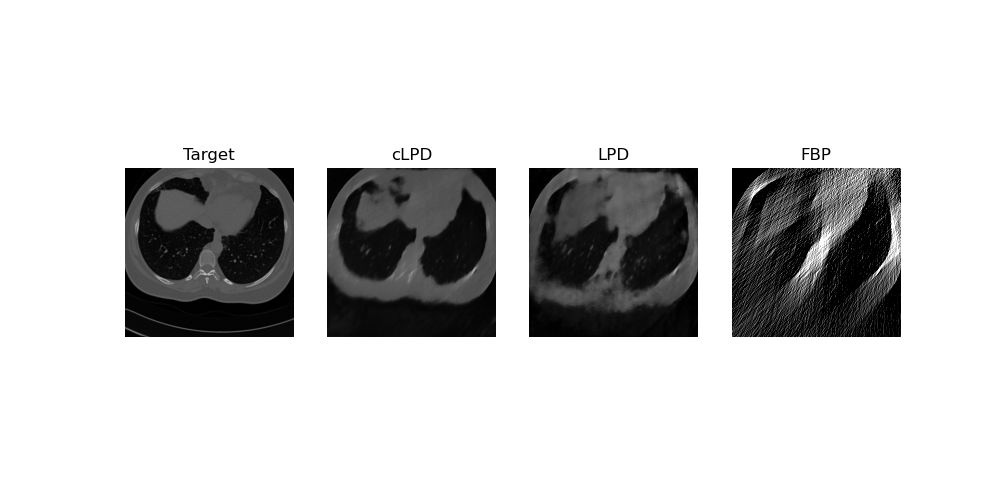}\label{fig:limited_extreme_low}}
    \caption{Visual results for a randomly picked image of the test set for all experimental settings. We highlight the results of our cLPD, standard LPD, FBP and the target reconstruction from left to right. With increasing noise levels, our approach (cLPD) is able to outperform the LPD more and more significantly. Both cLPD and LPD outperform FBP in all experimental settings. In the case of a limited angle geometry, cLPD reconstructs artifact free results while the standard LPD starts to blur. For high noise levels and the restricted setting especially, FBP is unsuitable as it introduces artifacts. }
    \label{fig:main}
\end{figure*}

This section details the results of the experiments. \newline
\textbf{1. Clinical setting}: For the clinical setting, the continuous learned primal dual performs on par with the classical learned primal dual algorithm. The structural similarity index measure (SSIM) for the standard LPD algorithm is slightly higher than for the continuous version while in terms of the peak signal-to-noise ratio (PSNR) cLPD outperforms LPD. The cLPD and LPD perform significantly better than FBP, both in terms of image quality metrics as well as visual results (see Subfigure~\ref{fig:full_low}).\newline
\textbf{2. Reduced dose setting}: To analyse the effect that a reduced dose has on the proposed algorithm, we additionally test an extreme low dose and sparse angle geometry.
    \begin{enumerate}
        \item[a)] \textit{Extreme low dose, full circle}: When decreasing the dose while measuring over the full circle, similarly as for the clinical setting, cLPD and LPD perform on par, while the average SSIM and PSNR decrease from $0.61$ to $0.58$ and $34$ to $32$, respectively. Comparing both algorithms to FBP, FBP is not able to handle the increased noise level (see visual results in Subfigure~\ref{fig:full_extreme_low}) while both cLPD and LPD reconstruct denoised images.
        \item[b)] \textit{Sparse angle, clinical dose}: Reducing the number of angles to measure from while keeping the X-ray dose at a clinical dose, the continuous version of the learned primal dual outperforms the classical LPD and FBP both in terms of SSIM and PSNR (see Table \ref{tab:ssim_psnr}). Visually, the FBP is not able to reconstruct any details of the image while cLPD and LPD are able to preserve most of the features.
        \item[c)] \textit{Sparse angle, extreme low dose}: Further reducing the dose by decreasing the X-ray dose and the number of angles to measure on, cLPD outperforms both the classical learned primal dual and FBP algorithm in terms of SSIM, PSNR and visual results. Whith increasing amounts of noise, the reconstructions of both cLPD and LPD get more blury and less detailed while the FBP algorithm produces noisy results without any high-level features.
    \end{enumerate}
\textbf{3. Restricted setting}: Analysing a restricted setting, we obtain the following results:
    \begin{enumerate}
        \item[a)] \textit{Limited angle, clinical dose}: Firstly testing on a clinical dose, in the restricted setting our proposed continuous learned primal dual outperforms the classical learned primal dual and FBP to an even greater extend. While the average SSIM and PSNR of the reconstructions produced by cLPD compared to 2.c) dropped by $0.09$ and $5.26$, respectively, the average SSIM and PSNR of the LPD reconstructions decreased by $0.13$ and $7.01$, respectively -- highlighting the robustness of the cLPD algorithm to noise. The visual results highlighted in Subfigure \ref{fig:limited_low} further highlight these advantages of the cLPD. Even for a restricted setting our method is able to preserve low-level features like the shape of the lungs and introducing barely any artifcats. The LPD and FBP algorithm however reconstruct artifact heavy images that do not resemble the target reconstructions.
        \item[b)] \textit{Limited angle, extreme low dose}: Secondly testing on an extreme low dose, the performance gap between our cLPD algorithm and both the standard LPD and FBP persists. Similiarly to the previous setting, the visual results (see Subfigure~\ref{fig:limited_extreme_low}) highlight the robustness to noise of the continuous version of the learned primal dual algorithm. 
    \end{enumerate}

\begin{landscape}
\begin{table*}[t]
    \centering
    \caption{Overview of mean structural similarity index measure (SSIM), peak signal-to-noise ration (PSNR) and their standard deviations for the experimental settings highlighted in Section \ref{sec:experimental_setup}. For experimental settings in which the noise level is comparatively low (1. and 2.a)), our proposed algorithm (cLPD) performs as well as its standard version (LPD). In these cases, both the cLPD and LPD outperform the classical FBP. With increasing noise levels, the advantages of NeuralODEs come into play and the cLPD outperforms both LPD and FBP (2.c)-3.b)).}
    \label{tab:ssim_psnr}
    \scalebox{1}{ 
    \begin{tabular}{l c c c} \toprule
    \textbf{Experimental setting} & \textbf{Algorithm} & \textbf{Mean SSIM ($\uparrow$)} & \textbf{Mean PSNR ($\uparrow$)} \\ \midrule
    \multirow{3}{*}{1.) Full angle, clinical dose } & cLPD & 0.6140 $\pm$ 0.1263 & \textbf{34.1787 $\pm$ 3.1489}\\
    & LPD & \textbf{0.6157 $\pm$ 0.1245} &34.1159 $\pm$ 3.1387\\
    & FBP &0.0602 $\pm$ 0.0207 & 16.8117 $\pm$ 1.8200\\\midrule
    \multirow{3}{*}{2.a) Full angle, extremely low dose} & cLPD & 0.5773 $\pm$ 0.1287 & \textbf{32.3713 $\pm$ 2.5793}\\
    & LPD & \textbf{0.5790 $\pm$ 0.1251} & 32.2299 $\pm$ 2.5228\\
    & FBP & 0.0213 $\pm$ 0.0086 & 11.2341 $\pm$ 1.8579\\\midrule
    \multirow{3}{*}{2.b) Sparse angle, clinical dose} & cLPD & \textbf{0.5627 $\pm$ 0.1269} & \textbf{31.2625 $\pm$ 2.2977}\\
    & LPD & 0.5571 $\pm$ 0.1169 & 30.8406 $\pm$ 2.1796\\
    & FBP & 0.0108 $\pm$ 0.0044 & 8.1548 $\pm$ 1.7939\\\midrule
    \multirow{3}{*}{2.c) Sparse angle, extremely low dose} & cLPD & \textbf{0.5316 $\pm$ 0.1232} & \textbf{29.6664 $\pm$ 1.9520}\\
    & LPD & 0.5265 $\pm$ 0.1185 & 29.2588 $\pm$ 1.8851\\ 
     & FBP & 0.0024 $\pm$ 0.0012 & 2.6769 $\pm$ 1.8622\\\midrule
    \multirow{3}{*}{3.a) Limited angle, clinical dose} & cLPD & \textbf{0.4465 $\pm$ 0.1099} & \textbf{24.4042 $\pm$ 1.6947}\\
    & LPD & 0.3951 $\pm$ 0.0937 & 22.4654 $\pm$ 1.7108\\
     & FBP & 0.0103 $\pm$ 0.0045 & 7.7079 $\pm$ 1.8264\\\midrule
    \multirow{3}{*}{3.b) Limited angle, extremely low dose} & cLPD & \textbf{0.4371 $\pm$ 0.1081} & \textbf{24.0181 $\pm$ 1.6651} \\
    & LPD & 0.3823 $\pm$ 0.0933 & 22.2501 $\pm$ 1.6938\\
     & FBP & 0.0037 $\pm$ 0.0018 & 3.4242 $\pm$ 2.0003\\
 \bottomrule
    \end{tabular}
    }
\end{table*}
\end{landscape}
\section{Discussion and Conclusions}
In this work, we introduced a continuous version of the learned primal dual algorithm for CT reconstruction. We showed that for a clinical, i.e., low noise setting, our approach performs as good as the vanilla learned primal dual. The more reduced the dose, i.e., the noisier the measurements, the bigger the gap in performance between cLPD and LPD gets with cLPD outperforming the discrete LPD. In comparison to FBP, cLPD significantly outperforms FBP in all experimental settings tested. Our approach has furthermore shown to be especially powerful for restricted settings with a limited angle geometry. In contrast to both LPD and FBP which fail to reconstruct any features, cLPD did not introduce artifacts. As NeuralODEs are provably robust to noise, introducing continuous blocks into the standard LPD algorithm showed to be successful in the experiments conducted. 

Interestingly, continuous LPD requires normalisation at every layer for a stable training whereas the standard LPD achieves best results without any form of normalisation. It would be interesting to further investigate the reasons for this difference. As cLPD uses the adjoint method, i.e., solving an ODE going backward in time, for backpropagation, normalisation might be required to stabilise the backward pass. 

Future work also includes exploring continous representations for other algorithms in CT reconstruction and inverse problems in general.

\bibliographystyle{IEEEbib}
\bibliography{strings,refs}

\end{document}